\documentclass[opre,nonblindrev]{informs3_homepage}

\OneAndAHalfSpacedXI
\usepackage{microtype}
\usepackage{graphicx}
\usepackage{booktabs}
\usepackage{xcolor}
\usepackage{array}
\usepackage[hidelinks]{hyperref}
\usepackage{suffix}

\usepackage{caption}
\usepackage{subcaption}

\usepackage{algorithm}
\usepackage{algorithmic}
\usepackage{tikz}
\usetikzlibrary{calc,backgrounds,arrows.meta}

\usepackage{enumitem}
\usepackage{textgreek}

\usepackage[informs]{autoconfig}

\DeclareMathAlphabet\mathbfcal{OMS}{cmsy}{b}{n}

\def\ProbSpace{\Omega}

\def\LinTS{LinTS}

\def\Policy{\pi}
\def\PolicyTS{\Policy^{\operatorname{\LinTS}}}

\def\Noise{\varepsilon}
\def\NoiseSD{\sigma}

\def\Arm{A}

\def\ArmSet{\mathcal{\Arm}}

\def\ChosenArm{\widetilde{\Arm}}
\def\OptimalArm{\Arm^\star}

\def\BayesRegret{\operatorname{BayesRegret}}
\def\History{\mathcal{F}}
\def\HistoryPlus{\History}

\def\Param{\Theta^\star}

\def\GainRate{\mathsf{G}}
\WithSuffix\def\GainRate*[#1]{\GainRate_{#1}}

\def\GapProb{\mathsf{q}}
\WithSuffix\def\GapProb*[#1]{\GapProb_{#1}}
\def\Deviation{\mathsf{D}}
\WithSuffix\def\Deviation*[#1]{\Deviation_{#1}}

\def\semdefleq{\preccurlyeq}
\def\semdefgeq{\succcurlyeq}

\def\TsSample{\widetilde{\Theta}}

\def\RewardValue{Y}

\def\Eye{\mathbb{I}}
\def\Dim{d}
\def\CovMatrix{\mathbf{\Gamma}}
\def\SymCovMatrix{\mathbf{\Sigma}}
\def\Estimator{\widehat{\Theta}}

\definecolor{darkpastelgreen}{rgb}{0.01, 0.75, 0.24}
\definecolor{cadmiumgreen}{rgb}{0.0, 0.42, 0.24}
\definecolor{airforceblue}{rgb}{0.0,0.1,0.8}

\def\Revised{}

\usepackage{natbib}
 \bibpunct[, ]{(}{)}{,}{a}{}{,}%
 \def\bibfont{\small}%
\TheoremsNumberedThrough     % Preferred (Theorem 1, Lemma 1, Theorem 2)
\ECRepeatTheorems

\EquationsNumberedBySection % (1.1), (1.2), ...

\MANUSCRIPTNO{MS-0001-1922.65}

\begin{document}
%%%%%%%%%%%%%%%%

%\setcounter{page}{1}
%\DOI{10.1287/xxxx.0000.0000}%


\RUNTITLE{
A General Elliptical Potential Lemma}

% \TITLE{Bundling Information Goods of Decreasing Value}
\TITLE{
The Elliptical Potential Lemma {\Revised for General Distributions} with an Application to Linear Thompson Sampling
}


\ARTICLEAUTHORS{%
	\AUTHOR{Nima Hamidi}
	\AFF{Department of Statistics, Stanford University, \EMAIL{hamidi@stanford.edu}}
	\AUTHOR{Mohsen Bayati}
	\AFF{
		Graduate School of Business, Stanford University, \EMAIL{bayati@stanford.edu}}
	} % end of the block

\ABSTRACT{%
In this note, we introduce a {\Revised general}  version of the well-known \emph{elliptical potential lemma} that is a widely used technique in the analysis of algorithms in sequential learning and decision-making problems. We consider a stochastic linear bandit setting where a decision-maker sequentially chooses among a set of given actions, observes their noisy rewards, and aims to maximize her cumulative expected reward over a decision-making horizon. The elliptical potential lemma is a key tool for quantifying uncertainty in estimating parameters of the reward function, but it requires the noise and the prior distributions to be Gaussian.  Our {\Revised general}  elliptical potential lemma relaxes this Gaussian requirement which is a highly non-trivial extension for a number of reasons; unlike the Gaussian case, there is no closed-form solution for the covariance matrix of the posterior distribution, the covariance matrix is not a deterministic function of the actions, and  the covariance matrix is not decreasing with respect to the semidefinite inequality. While this result is of broad interest, we showcase an application of it to prove an improved Bayesian regret bound for the well-known Thompson sampling algorithm in stochastic linear bandits with changing action sets where prior and noise distributions are general. This bound is minimax optimal up to constants.
}%


% \HISTORY{This paper was first submitted on April 12, 1922 and has been with the authors for 83 years for 65 revisions.}

\maketitle
%%%%%%%%%%%%%%%%%%%%%%%%%%%%%%%%%%%%%%%%%%%%%%%%%%%%%%%%%%%%%%%%%%%%%%
%\newpage

\section{Introduction}
\label{sec:intro}
In sequential linear prediction problems, the classical elliptical potential lemma is a key technique to quantify the decrease in the uncertainty of the model as more observations are obtained. This lemma was first introduced by \citet{lai1982least} to analyze stochastic regression
and was later applied to study the sequential ridge forecaster \citep{cesa2006prediction} and in proving regret bounds for variants of the stochastic linear bandit problem \citep{auer2002using,dani2008stochastic,chu2011contextual,abbasi2011improved,agrawal2013thompson,li2019nearly}. To state the elliptical potential lemma, let $\Arm_1,\Arm_2,\cdots$ be a sequence of vectors in $\IR^\Dim$ that satisfy $\Norm{\Arm_t}_2\leq1$ for all $t\geq1$. For a fixed constant $\lambda$ with $\lambda\geq1$, define the sequence of matrices $\{\SymCovMatrix_t\}_{t\ge0}$ as follows:
\[
   \SymCovMatrix_1^{-1}:=\lambda\Eye_\Dim~~~,~~~ \SymCovMatrix_t^{-1}
    :=
    \lambda\Eye_\Dim
    +
    \sum_{\tau=1}^{t-1}\Arm_\tau\Arm_\tau^\top\,.
\]
The \emph{elliptical potential lemma} then asserts that
\begin{align}
    \sum_{t=1}^{T}\Arm_t^\top\SymCovMatrix_{t}\Arm_t
    \leq
    2\log\frac{\det\SymCovMatrix_1}{\det\SymCovMatrix_{T+1}}
    \leq
    2\Dim\log\left(1+\frac{T}{\lambda\Dim}\right).
    \label{eq:classical-epl}
\end{align}
Recently, \citet{carpentier2020elliptical} presented a new proof for this inequality that additionally yields similar bounds for $\sum_{t=1}^{T}\Arm_t^\top\SymCovMatrix_t^p\Arm_t$ for any $p>0$.

In this paper, we generalize \eqref{eq:classical-epl} from a different perspective. Notice that, intuitively, $\SymCovMatrix_t$ captures how much information is available in each direction in a linear model. Specifically, let $\Param$ be sampled from $\Normal{0,\lambda^{-1}\Eye_\Dim}$ and assume that, for each $t\geq1$, an outcome $\RewardValue_t=\Inner{\Arm_t,\Param}+\Noise_t$ is observed where $\Noise_t$ is a standard Gaussian noise, independent of the past. It is well-known that $\SymCovMatrix_t$ is the covariance matrix of the posterior distribution of $\Param$ conditional on the data available up to time $t-1$, namely, $\Arm_1,\RewardValue_1,\ldots,\Arm_{t-1},\RewardValue_{t-1}$. 

The primary contribution of this note is to generalize the elliptical potential bound in \eqref{eq:classical-epl} to \emph{any arbitrary prior and noise distributions}. This generalization is non-trivial, compared to the Gaussian case, for a number of reasons. First,  unlike the Gaussian case, there is no closed form solution for the covariance matrix of the posterior distribution. Second, this covariance matrix is a deterministic function of $\Arm_1,\Arm_2,\ldots,\Arm_{t-1}$ in the Gaussian case but in general it is a function of the whole history $\Arm_1,\RewardValue_1,\Arm_2,\RewardValue_2,\ldots,\Arm_{t-1},\RewardValue_{t-1},\Arm_t$. Third, the covariance matrix of the posterior distribution for the Gaussian case is non-increasing with respect to semidefinite inequality (i.e.,
$\SymCovMatrix_1\semdefgeq\SymCovMatrix_2\semdefgeq\cdots$) but this property breaks down in the general case. {\Revised Because of the first two reasons, the covariance update equation
$\SymCovMatrix_t^{-1}=\lambda\Eye_\Dim	+	\sum_{\tau=1}^{t-1}\Arm_\tau\Arm_\tau^\top$ is incorrect
for the posterior covariance for general distributions. However,  \cref{eq:classical-epl} still holds as an algebraic inequality, for example see  \citep{dani2008stochastic,abbasi2011improved}. In contrast, our result is for an updated version of  \cref{eq:classical-epl} that reflects the true covariance matrices.}

The secondary contribution of this note is to showcase an application of the aforementioned generalization of the elliptical potential lemma in combination with the proof techniques in \citep{dong2018information,kalkanli2020improved} to prove an $\Order{\Dim\sqrt{T\log T}}$ bound for the Bayesian regret of the well-known linear Thompson sampling (LinTS) algorithm. This result is proved under mild distributional assumptions and allows the action sets to change at each round. This result extends the regret bound of \citet{dong2018information} as they require action sets to be fixed (which excludes for example the $k$-armed contextual bandit problem). Our result also generalizes the bound of \cite{kalkanli2020improved} by relaxing the Gaussian assumption. {\Revised We note that the above comparison is only made for results that provide the tightest regret bound of  $\Order{\Dim\sqrt{T\log T}}$. In fact, \citet{russo2014learning} study LinTS with changing action sets, general bounded prior, and sub-Gaussian noise distributions. They prove a Bayesian regret bound of $\Order{\Dim\log T\sqrt{T}}$ which is worse than our regret bound and the bounds of \citet{dong2018information,kalkanli2020improved} by a factor of $\sqrt{\log T}$.}

Our {\Revised general}  elliptical potential lemma is presented in \cref{sec:potential} and its application to the Bayesian regret of LinTS is provided in \cref{sec:bayes-ts}. Proofs are deferred to \crefrange{sec:potential-proofs}{sec:bayes-ts-proofs}.
\section{Elliptical Potential {\Revised for General Distributions}}
\label{sec:potential}
In this section, we present our main result. Let $(\ProbSpace,\History,\ProbSymb)$ be a probability space and $\History_1\subseteq\History_2\subseteq\cdots\subseteq\History$ be an increasing sequence of $\sigma$-algebras that are meant to encode the information available up to time $t$. Let $\Param:\ProbSpace\to\IR^\Dim$ be the true parameters vector and assume that $\Norm{\Param}_2\leq1$ almost surely. Furthermore, let $\Arm_1,\Arm_2,\cdots:\ProbSpace\to\IR^\Dim$ be a sequence of random vectors such that for all $t\geq1$, $\Arm_t$ is $\History_t$-measurable and $\Norm{\Arm_t}_2\leq1$ almost surely. More information about $\Param$ is then made available sequentially through a sequence of observations $\RewardValue_1,\RewardValue_2,\cdots:\ProbSpace\to\IR$ where $\RewardValue_t$ is $\HistoryPlus_{t+1}$-measurable and
\begin{align*}
    \Expect*{\RewardValue_t\Given\History_t,\Param}
    =
    \Inner{\Param,\Arm_t}
    ~~~~~~~~\text{and}~~~~~~~~
    \Var*{\RewardValue_t\Given\History_t,\Param}
    \leq
    \NoiseSD^2,
\end{align*}
for all $t\geq1$ almost surely. We denote the posterior covariance matrix of $\Param$ at time $t$ by $\CovMatrix_t$, that is
\begin{align*}
    \CovMatrix_t
    :=
    \Var{\Param\Given\History_t}.
\end{align*}
It follows from the definition that $\CovMatrix_t$ is a stochastic positive semi-definite matrix in $\IR^{\Dim\times\Dim}$ that is $\HistoryPlus_t$-adapted. Notice that, nonetheless, it is \emph{not} true in general that $\CovMatrix_{t+1}\SemDefLeq\CovMatrix_t$. To see this, let $\Dim=1$ and $\Param\in\{0,1/4,3/4\}$ be such that the prior distribution of $\Param$ satisfies $\Prob{\Param=1/4}=3p$ and $\Prob{\Param=3/4}=p$ for some small $p>0$. Also, define $\Arm_t:=1$ for all $t\geq1$ and assume $\RewardValue_t$ is a Bernoulli random variable with mean $\Param$. We further let $\History_t$ be the smallest $\sigma$-algebra generated by $\RewardValue_1,\cdots,\RewardValue_{t-1}$. In this case, it is easy to see that $\CovMatrix_1=\Var{\Param\Given\History_1}=\Var{\Param}$ can be made arbitrarily small by choosing a sufficiently small $p>0$. In this case, notice that, whenever $\RewardValue_1=1$, the distribution of $\Param$ conditional on $\History_2$ is uniform over $\{1/4,3/4\}$ which gives us $\CovMatrix_2=1/4>\CovMatrix_1$. This can be shown by noting that $\Prob{\Param=0\Given\RewardValue_1=1}=0$ and
\begin{align*}
    \frac{\Prob{\Param=\frac14\Given\RewardValue_1=1}}{\Prob{\Param=\frac34\Given\RewardValue_1=1}}
    =
    \frac
    {\Prob{\Param=\frac14}\cdot\Prob{\RewardValue_1=1\Given\Param=\frac14}}
    {\Prob{\Param=\frac34}\cdot\Prob{\RewardValue_1=1\Given\Param=\frac34}}
    =
    \frac
    {3p\cdot\frac14}
    {p\cdot\frac34}
    =
    1.
\end{align*}
We can, however, apply the law of total variance to get
\begin{align*}
    \Expect[\big]{\CovMatrix_{t+1}\Given\History_t}
    \SemDefLeq
    \Expect[\big]{\CovMatrix_{t+1}\Given\History_t}
    +
    \Var[\big]{\Expect{\Param\Given\History_{t+1}}\Given\History_t}
    =
    \Var[\big]{\Param\Given\History_t}
    =
    \CovMatrix_t.
\end{align*}
This inequality only shows that $\CovMatrix_t$ decreases in expectation but does not tell us \emph{how much} the expected variance decreases at each round. The next lemma provides a stronger bound. The proofs of this lemma and other results of this section are postponed to \cref{sec:potential-proofs}.
\begin{lem}[Stochastic variance reduction]
\label{lem:var-reduction}
Whenever the above-mentioned assumptions hold, for all $t\geq1$, we have
\begin{align*}
    \Expect{\CovMatrix_{t+1}\Given\History_t}
    \SemDefLeq
    \CovMatrix_t-\frac{\CovMatrix_t^\top \Arm_t \Arm_t^\top\CovMatrix_t}{\NoiseSD^2+\Arm_t^\top\CovMatrix_t\Arm_t}
\end{align*}
almost surely.
\end{lem}
Lemma \ref{lem:var-reduction} demonstrates that the posterior covariance decays \emph{in expectation}. As we discussed by the above example, this does not necessarily hold for $\CovMatrix_t$ almost surely. In fact, this is the most challenging roadblock in establishing a general version of the elliptical potential lemma as one can increase $\sum_{t=1}^{T}{\Arm_t}^\top\CovMatrix_t\Arm_t$ by defining $\Arm_t$'s adaptively, to be aligned with high variance directions. The following lemma which is the main technical contribution of this note introduces a methodology to overcome this difficulty.
\def\PsdMatrix{\boldsymbol{\Sigma}}
\def\PsdCone{\mathbf{S}_{+}^\Dim}
\def\PsdIncr{\boldsymbol{\Lambda}}
\begin{lem}
\label{lem:logdet-properties}
For $x>0$ and positive semi-definite matrix $\PsdMatrix$, define $f(\PsdMatrix,x)=\log\det(\Eye+x\PsdMatrix)$. Then, $f(\cdot,\cdot)$ satisfies the following properties:
\begin{enumerate}
\item For any fixed $x>0$, $f(\cdot,x)$ is a concave function on the positive semi-definite cone.
\item If $\PsdMatrix$ is an invertible and positive semidefinite matrix then $f(\PsdMatrix,x)$  satisfies the following variational representation
\begin{align}
    f(\PsdMatrix,x)
    &=
    \log\det\left(\PsdMatrix^{\frac12}(\PsdMatrix^{-1}+x\Eye)\PsdMatrix^{\frac12}\right)\nonumber\\
    &=
    \sup_{\PsdIncr\semdefleq x\Eye}
    \log\det\left(\PsdMatrix^{\frac12}(\PsdMatrix^{-1}+\PsdIncr)\PsdMatrix^{\frac12}\right).
    \label{eq:logdet-var-repr}
\end{align}
\item For any vector $V\in\IR^\Dim$, we have
\begin{align}
    \log(1+V^\top\PsdMatrix V)
    +
    f(\PsdMatrix',x)
    \leq
    f(\PsdMatrix,x+V^\top V)
    \label{eq:logdet-incr}
\end{align}
where
$\PsdMatrix'
:=
\PsdMatrix-\frac{\PsdMatrix VV^\top\PsdMatrix}{1+V^\top\PsdMatrix V}
=
\PsdMatrix^{\frac12}\left(\Eye-\frac{\PsdMatrix^{\frac12}VV^\top\PsdMatrix^{\frac12}}{1+V^\top\PsdMatrix V}\right)\PsdMatrix^{\frac12}
$.
\end{enumerate}
\end{lem}
Using this result, we are now ready to conclude this section by stating our elliptical potential inequality for general distributions.
\begin{thm}[Elliptical Potential {\Revised for General Distributions}]
\label{thm:stochastic-elliptical}
Under the above assumptions, the following inequality holds,
\begin{align*}
    \Expect*{\sum_{t=1}^{T}
    {\Arm_t}^\top\CovMatrix_t\Arm_t
    }
    \leq
    2\max(\NoiseSD^2,1)\log\det(\Eye+T\,\CovMatrix_1)\,.
\end{align*}
\end{thm}
\section{Linear Thompson Sampling}
\label{sec:bayes-ts}
In this section, we apply \cref{thm:stochastic-elliptical} to show that, up to constants, Linear Thompson Sampling (\LinTS{}) achieves an optimal prior-independent Bayesian regret. This statement is stronger than the bound in \citep{dong2018information} as it allows for changing action sets and it is more general than \citep{kalkanli2020improved} since it does not require Gaussian assumption for the prior and noise distributions. {\Revised As noted before, \citet{russo2014learning} also study LinTS with changing action sets and without Gaussian assumptions for the prior or noise distributions, but their Bayesian regret bound is not optimal for this class of problems. Specifically, there is an additional  $\sqrt{\log T}$ factor in their bound compared to the one we provide here.}

\begin{algorithm}[t]
\caption{Linear Thompson sampling (\LinTS{})}
\label{alg:lin-ts}
\begin{algorithmic}[1]
\FOR{$t=1,2,\cdots$}
\STATE Observe the actions set $\ArmSet_t\subseteq\IR^\Dim$.
\STATE Sample $\TsSample_t\sim\Prob{\Param\Given\History_t}$.
\STATE $\ChosenArm_t\gets \Argmax_{\Arm\in\ArmSet_t}\Inner[\big]{\Arm,\TsSample_t}$
\STATE Observe reward $\RewardValue_t$.
\ENDFOR
\end{algorithmic}
\end{algorithm}

First, let $\Estimator_t$ be the posterior mean of $\Param$ at time $t$. We also denote by $\OptimalArm_t$ and $\ChosenArm_t$ the optimal arm and the selected arm at time $t$ respectively. Now notice that the expected regret at time $t$ can be expressed as $\Expect{(\Param-\Estimator_t)^\top\OptimalArm_t}$. In order to bound this, we utilize the idea in the proof of Proposition 5 in \citep{russo2016information} which was later generalized by \cite{kalkanli2020improved}. This idea avoids constructing confidence sets around $\Param-\Estimator_t$ that introduce an additional $\sqrt{\log T}$ term. We bring this idea and a slightly modified proof for that here. The proof of all results in this section is deferred to \cref{sec:bayes-ts-proofs}.
\begin{lem}
\label{lem:double-cauchy}
Let $X,Z$ be two random vectors in $\IR^\Dim$. Then, we have
\begin{align*}
    \Expect*{X^\top Z}^2
    \leq
    \Dim\Trace*{\Expect*{XX^\top}\Expect*{ZZ^\top}}.
\end{align*}
\end{lem}
Notice that this lemma does not require independence between $X$ and $Z$. Therefore, one can set $X:=\Param-\Estimator_t$ and $Z:=\OptimalArm_t$. Then, the main step in the proof is observing that $\Expect{XX^\top\Given\History_t}=\CovMatrix_t$, $\Expect{ZZ^\top\Given\History_t}=\Expect{\ChosenArm_t\ChosenArm_t^\top\Given\History_t}$, and $X$ and $\ChosenArm_t$ are independent conditional on $\History_t$.
\begin{thm}
\label{thm:ts-bayes-regret}
Let $\Param$ be such that $\Norm{\Param}_2\leq1$ almost surely and $\History_t$ be the $\sigma$-algebra generated by $(\ArmSet_1,\ChosenArm_1,\RewardValue_1,\ArmSet_2,\cdots,\ArmSet_t,\ChosenArm_t)$. Furthermore, assume that
\begin{align*}
    \Expect*{\RewardValue_t\Given\History_t,\Param}
    =
    \Inner{\Param,\ChosenArm_t}
    ~~~~~~~~\text{and}~~~~~~~~
    \Var*{\RewardValue_t\Given\History_t,\Param}
    \leq
    \NoiseSD^2,
\end{align*}
almost surely. Then, the following regret bound holds for \LinTS{} (\cref{alg:lin-ts}) when it has access to the true prior and noise distributions:
\begin{align}
    \BayesRegret(T,\PolicyTS)
    \leq
    \sqrt{2\max(\NoiseSD^2,1)\Dim T\log\det\left(1+T\CovMatrix_1\right)}.
    \label{eq:bayes-ts-bound}
\end{align}
\end{thm}
\begin{rem}
The assumption that $\Norm{\Param}_2\leq1$ almost surely implies that $\CovMatrix_1\semdefleq\Eye$. Hence, we have the trivial bound $\log\det(1+T\CovMatrix_1)\leq\Dim\log(1+T)$ which in turn leads to
\begin{align*}
    \BayesRegret(T,\PolicyTS)
    \leq
    \Dim\sqrt{2\max(\NoiseSD^2,1)T\log\left(1+T\right)}.
\end{align*}
\end{rem}
\begin{rem}
As shown in \citep{hamidi2020worst}, the assumption that \LinTS{} uses the true prior distribution for $\Param$ is crucial, as the Bayesian regret of \LinTS{} can grow linearly for $\exp(C\Dim)$ rounds for some constant $C>0$ under a mild distributional mismatch.
\end{rem}
\begin{rem}
An interesting aspect of this result is that it does not require the noise to be bounded or sub-Gaussian. Having a bounded second moment suffices for \cref{eq:bayes-ts-bound} to hold. {\Revised For the special case of $k$-armed (and non-contextual) bandits, \cite{bubeck2013bandits} show that when noise has a bounded second moment one can obtain matching regret bounds as when noise is sub-Gaussian. It is an open question whether their proof technique can be adapted to the setting we study here, without extending \cref{eq:classical-epl}. Moreover, \cite{bubeck2013bandits} use a UCB type algorithm with a modified mean reward estimator based on robust statistics. It is intriguing that \cref{{thm:ts-bayes-regret}} does not require modifying LinTS.}
\end{rem}

\ACKNOWLEDGMENT{The authors gratefully acknowledge an insightful suggestion by Ofer Zeitouni. This work was supported by the National Science Foundation award CMMI: 1554140.}

\bibliography{papers,mypapers,books}
\bibliographystyle{plainnat}

\begin{APPENDICES}
\section{Proof of \cref{sec:potential}}
\label{sec:potential-proofs}
\begin{proof}[Proof of \cref{lem:var-reduction}]
\def\Vector{V}
\NewExpect\Expectt{\ExpectSymb_t}
Let $\Expectt{\cdot}$ be the shorthand for $\Expect{\cdot\Given\History_t}$.
First, we prove the claim for $\Param$ with $\Expectt{\Param}=0$. It suffices to prove that
\begin{align*}
    \Vector^\top\Expectt{\CovMatrix_{t+1}}\Vector
    =
    \Expectt*{\Var[\Big]{\Inner{\Param,\Vector}\Given\History_{t+1}}}
    \leq
    \Vector^\top\CovMatrix_t\Vector-\frac{\big(\Arm_t^\top\CovMatrix_t\Vector\big)^2}{\NoiseSD^2+\Arm_t^\top\CovMatrix_t\Arm_t}
\end{align*}
for any fixed vector $\Vector\in\IR^\Dim$. Denoting by $\History_t^{A}$ the set of $\History_{t}$-adaptable random variables, we have
\begin{align}
    \Expectt*{\Var[\Big]{\Inner{\Param,\Vector}\Given\History_{t+1}}}
    &=
    \Expectt*{
        \inf_{W\in\History_{t+1}^{A}}
        \Expect*{\Big(\Inner{\Param,\Vector}-W\Big)^2\Given\History_{t+1}}
    }\nonumber\\
    &\leq
    \Expectt*{
        \inf_{a\in\IR}
        \Expect*{\Big(\Inner{\Param,\Vector}-a\RewardValue_t\Big)^2\Given\History_{t+1}}
    }\nonumber\\
    &\leq
    \inf_{a\in\IR}
    \Expectt*{
        \Expect*{\big(\Inner{\Param,\Vector}-a\RewardValue_t\big)^2\Given\History_{t+1}}
    }\nonumber\\
    &=
    \inf_{a\in\IR}
    \Expectt*{
        \big(\Inner{\Param,\Vector}-a\RewardValue_t\big)^2
    }\nonumber\\
    &=
    \inf_{a\in\IR}\Big(
    \Expectt[\big]{\Inner{\Param,\Vector}^2}-2a\Expectt[\big]{\Inner{\Param,\Vector}\RewardValue_t}+a^2\Expectt[\big]{\RewardValue_t^2}\Big)\nonumber\\
    &=
    \Expectt[\big]{\Inner{\Param,\Vector}^2}-\frac{\Expectt[\big]{\Inner{\Param,\Vector}\RewardValue_t}^2}{\Expectt[\big]{\RewardValue_t^2}}\,.\label{eq:pf-lemma2.1-first-display-last-equation}
\end{align}
Next, we can simplify each of the two expectations on the right hand side of \cref{eq:pf-lemma2.1-first-display-last-equation}. For the first term, using the assumption $\Expectt{\Param}=0$, we have
\begin{align*}
    \Expectt[\big]{\Inner{\Param,\Vector}^2}
    =
    \Expectt[\big]{\Vector^\top{\Param}^\top\Param\Vector}
    =
    \Vector^\top\CovMatrix_t\Vector.
\end{align*}
For the second expectation, the numerator can also be computed in the following way
\begin{align*}
    \Expectt[\Big]{\Inner{\Param,\Vector}\RewardValue_t}
    %&=\Expectt[\Big]{\Expect[\big]{\RewardValue_t^2\Given\History_t,\Param}} \\
    &=
    \Expectt[\Big]{\Expect[\big]{\Inner{\Param,\Vector}\RewardValue_t\Given\History_t,\Param}}\\
    &=
    \Expectt[\Big]{\Inner{\Param,\Vector}\cdot\Expect[\big]{\RewardValue_t\Given\History_t,\Param}}\\
    &=
    \Expectt[\Big]{\Inner{\Param,\Vector}\Inner{\Param,\Arm_t}}\\
    &=
    \Expectt[\Big]{\Arm_t^\top{\Param}^\top\Param\Vector}
    =
    \Arm_t^\top\CovMatrix_t\Vector.
\end{align*}
Finally, for the denominator of the second expectation we have
\begin{align*}
    \Expectt[\big]{\RewardValue_t^2}
    &=
    \Expectt[\Big]{\Var[\big]{\RewardValue_t\Given\History_t,\Param}+\Expect[\big]{\RewardValue_t\Given\History_t,\Param}^2}\\
    &\leq
    \NoiseSD^2+\Expectt[\big]{\Inner{\Param,\Arm_t}^2}\\
    &=
    \NoiseSD^2+\Arm_t^\top\CovMatrix_t\Arm_t.
\end{align*}
By putting all the above together, we get the desired result.
\def\ParamZero{\mu^\star}
Finally, whenever $\Expectt{\Param}\neq0$, define $\ParamZero:=\Param-\Expectt{\Param}$ and $Z_t:=\RewardValue_t-\Inner{\Expectt{\Param},\Arm_t}$. Note that $\Var{\ParamZero\Given\History_t}=\CovMatrix_t$, $\Expect{Z_t\Given\History_t,\ParamZero}=\Inner{\ParamZero,\Arm_t}$ almost surely, and $\Expect{\Var{Z_t\Given\History_t,\ParamZero}}=\Expect{\Var{\RewardValue_t\Given\History_t,\Param}}\leq\NoiseSD^2$. Therefore, we can apply the result we just proved (for the case $\Expectt{\Param}=0$) to $\ParamZero$ and $Z_t$ and get
\begin{align*}
    \Expectt{\Var{\ParamZero\Given\History_{t+1}}}
    \SemDefLeq
    \CovMatrix_t-\frac{\CovMatrix_t^\top\Arm_t\Arm_t^\top\CovMatrix_t}{\NoiseSD^2+\Arm_t^\top\CovMatrix_t\Arm_t}\,.
\end{align*}
Combining this by the fact that $\Expectt{\CovMatrix_{t+1}}=\Expectt{\Var{\ParamZero\Given\History_{t+1}}}$, we conclude the result for $\Param$ and $\RewardValue_t$.
\end{proof}

\begin{proof}[Proof of \cref{lem:logdet-properties}]
The concavity of $f(\cdot,x)$ follows from the fact that $\log\det(\cdot)$ is concave over the positive semi-definite cone, see \cite[page 74]{boyd2004convex}, and $f(\cdot,x)$ is obtained by composing $\log\det(\cdot)$ with a linear function of $\PsdMatrix$. 

The variational representation can be obtained by noting that $\log\det(\cdot)$ is increasing with respect to the positive semi-definite order `$\semdefleq$'.

We now turn to proving \cref{eq:logdet-incr}. We first assume that $\PsdMatrix$ is invertible. In this case, we have $\PsdMatrix'^{-1}=\PsdMatrix^{-1}+VV^\top$, using Sherman–Morrison formula. From \cref{eq:logdet-var-repr} and using $\det(AB)=\det(A)\det(B)$, we get that
\begin{align*}
    f(\PsdMatrix,x+V^\top V)
    &=
    \sup_{\PsdIncr\semdefleq (x+V^\top V)\Eye}
    \log\det(\PsdMatrix^{\frac12}(\PsdMatrix^{-1}+\PsdIncr)\PsdMatrix^{\frac12})\\
    &\overset{(a)}{\geq}
    \sup_{\PsdIncr'\semdefleq x\Eye}
    \log\det(\PsdMatrix^{\frac12}(\PsdMatrix^{-1}+VV^\top+\PsdIncr')\PsdMatrix^{\frac12})\\
    &=
    \sup_{\PsdIncr'\semdefleq x\Eye}
    \log\det(\PsdMatrix^{\frac12}(\PsdMatrix'^{-1}+\PsdIncr')\PsdMatrix^{\frac12})\\
    &=
    \log\det(\PsdMatrix)+
    \log\det(\PsdMatrix'^{-1}+x\Eye)\\
    &=
    \log\det(\PsdMatrix')-\log\det\left(\Eye-\frac{\PsdMatrix^{\frac12}VV^\top\PsdMatrix^{\frac12}}{1+V^\top\PsdMatrix V}\right)+\log\det(\PsdMatrix'^{-1}+x\Eye)\\
    &\overset{(b)}{=}
    \log\det(\PsdMatrix')-\log\left(1-\frac{V^\top\PsdMatrix V}{1+V^\top\PsdMatrix V}\right)+\log\det(\PsdMatrix'^{-1}+x\Eye)\\
    &=
    \log\det(\PsdMatrix')-\log\left(\frac{1}{1+V^\top\PsdMatrix V}\right)+\log\det(\PsdMatrix'^{-1}+x\Eye)\\
    &=
    \log\det(\PsdMatrix')+\log\left(1+V^\top\PsdMatrix V\right)+\log\det(\PsdMatrix'^{-1}+x\Eye)\\
    &=
    \log\left(1+V^\top\PsdMatrix V\right)
    +
    \log\det\left(\PsdMatrix'^{\frac12}(\PsdMatrix'^{-1}+x\Eye)\PsdMatrix'^{\frac12}\right)\\
    &=
    \log\left(1+V^\top\PsdMatrix V\right)
    +
    f(\PsdMatrix',x).
\end{align*}
The inequality (a) uses the triangle inequality
\begin{align*}
    \NormOp{\PsdIncr'+V^\top V}\leq\NormOp{\PsdIncr'}+\NormOp{V^\top V}=\NormOp{\PsdIncr'}+VV^\top.
\end{align*}
and the equality (b) is obtained by observing that $\det(\Eye+ZZ^\top)=1+Z^\top Z$ for any vector $V$.

\def\Pert{\epsilon} % perturbation
It only remains to prove \cref{eq:logdet-incr} for a non-invertible matrix $\PsdMatrix$. In this case, for $\Pert>0$, we define $\PsdMatrix_\Pert=\PsdMatrix+\Pert\Eye$ and $\PsdMatrix'_\Pert
:=
\PsdMatrix_\Pert-\frac{\PsdMatrix_\Pert VV^\top\PsdMatrix_\Pert}{1+V^\top\PsdMatrix_\Pert V}$. Clearly, $\PsdMatrix_\Pert$ is invertible. Therefore, we can apply \cref{eq:logdet-incr} to $\PsdMatrix_\Pert$ to obtain
\begin{align*}
    \log(1+V^\top\PsdMatrix_\Pert V)
    +
    f(\PsdMatrix'_\Pert,x)
    \leq
    f(\PsdMatrix_\Pert,x+V^\top V).
\end{align*}
The claim then follows the continuity of the above expressions with respect to $\Pert$ on $[0,\infty]$.
\end{proof}

\begin{proof}[Proof of \cref{thm:stochastic-elliptical}]
Without loss of generality, we can assume that $\NoiseSD\leq 1$ to simplify the analysis. Otherwise, we can re-scale each action $\Arm_t$ and the noise by a factor $1/\max(\NoiseSD,1)$ and under this transformation the property $\Norm{\Arm_t}\leq1$ continues to hold.

Now, notice that since $\Norm{\Param}_2\leq1$ and $\Norm{\Arm_t}\leq1$ almost surely, we have ${\Arm_t}^\top\CovMatrix_t\Arm_t\leq1$ for all $t\in[T]$ almost surely. Next, the fact that $x\leq2\log(1+x)$ for all $x\in[0,1]$ implies that
\begin{align}
    {\Arm_t}^\top\CovMatrix_t\Arm_t
    &\leq
    2\log\left(1+{\Arm_t}^\top\CovMatrix_t\Arm_t\right).
    \label{eq:linear-to-log-ineq}
\end{align}

We now prove the main result inductively. For $T=1$, it suffices to note that
\begin{align*}
    \Eye+\CovMatrix_1^{\frac12}\Arm_1{\Arm_1}^\top\CovMatrix_1^{\frac12}
    \semdefleq
    \Eye+\CovMatrix_1^{\frac12}\,\Eye\,\CovMatrix_1^{\frac12}
    \semdefleq
    \Eye+\CovMatrix_1\,.
\end{align*}
For $T>1$, we can use the induction hypothesis for $T-1$ and get that
\begin{align*}
    \Expect*{\sum_{t=2}^{T}
    {\Arm_t}^\top\CovMatrix_t\Arm_t
    \Given
    \Arm_1,\RewardValue_1
    }
    \leq
    2\log\det(\Eye+(T-1)\CovMatrix_2)
\end{align*}
almost surely. Using the concavity of $\log\det(\cdot)$, it follows from Jensen's inequality and Lemma \ref{lem:var-reduction} that
\begin{align*}
    \Expect*{\sum_{t=2}^{T}
    {\Arm_t}^\top\CovMatrix_t\Arm_t
    \Given
    \Arm_1
    }
    &=
    \Expect*{
    \Expect*{
    \sum_{t=2}^{T}
    {\Arm_t}^\top\CovMatrix_t\Arm_t
    \Given
    \Arm_1, \RewardValue_1
    }
    \Given
    \Arm_1
    }\nonumber\\
    &\leq
    \Expect[\Big]{2\log\det(\Eye+(T-1)\CovMatrix_2)\Given\Arm_1}\nonumber\\
    &\leq
    2\log\det\left(\Eye+(T-1)\Expect*{\CovMatrix_2\Given\Arm_1}\right)\nonumber\\
    &\leq
    2\log\det\left(\Eye+(T-1)\left(\CovMatrix_1-\frac{\CovMatrix_1^\top\Arm_1 \Arm_1^\top\CovMatrix_1}{1+\Arm_1^\top\CovMatrix_1\Arm_1}\right)\right)\\
    &\leq
    2f\left(\CovMatrix_1',T-1\right).
\end{align*}
where $\CovMatrix_1':=\CovMatrix_1-\frac{\CovMatrix_1^\top\Arm_1 \Arm_1^\top\CovMatrix_1}{1+\Arm_1^\top\CovMatrix_1\Arm_1}$.
Finally, we apply \cref{eq:logdet-incr} in \cref{lem:logdet-properties} and \cref{eq:linear-to-log-ineq} to get that
\begin{align*}
    \Expect*{\sum_{t=1}^{T}
    {\Arm_t}^\top\CovMatrix_t\Arm_t
    }
    &\leq
    2\,\Expect*{
    \log\left(1+{\Arm_1}^\top\CovMatrix_1\Arm_1\right)
    +
    f\left(\CovMatrix_1',T-1\right)
    }\\
    &\leq
    2\,\Expect*{
    f\left(\CovMatrix_1,T\right)
    }\\
    &=
    2\log\det\left(\Eye+T\CovMatrix_1\right).
\end{align*}
\end{proof}
\section{Proofs of \cref{sec:bayes-ts}}
\label{sec:bayes-ts-proofs}
\begin{proof}[Proof of \cref{lem:double-cauchy}]
\def\Basis{U}
First, we observe that for any unitary matrix $\Basis$, if one defines $X':=\Basis X$ and $Z':=\Basis Z$, we have that
\begin{align*}
    \Expect*{X^\top Z}=\Expect*{X'^\top Z'}
\end{align*}
and
\begin{align*}
    \Trace*{\Expect*{X'X'^\top}\Expect*{Z'Z'^\top}}
    &=
    \Trace*{\Expect*{\Basis XX^\top\Basis^\top}\Expect*{\Basis ZZ^\top\Basis^\top}}\\
    &=
    \Trace*{\Basis\Expect*{XX^\top}\Basis^\top\Basis \Expect*{ZZ^\top}\Basis^\top}\\
    &=
    \Trace*{\Basis^\top\Basis\Expect*{XX^\top}\Basis^\top \Basis\Expect*{ZZ^\top}}\\
    &=
    \Trace*{\Expect*{XX^\top}\Expect*{ZZ^\top}}.
\end{align*}
These equalities imply that it suffices to prove the statement for $(X',Z')$ instead of $(X,Z)$. Now we choose $\Basis$ so that $\Expect*{Z'Z'^\top}=\Basis\Expect*{ZZ^\top}\Basis^\top$ is diagonal. This can be done through the singular value decomposition of $\Expect*{ZZ^\top}$. Then, notice that
\begin{align*}
    \Dim\Trace*{\Expect*{X'X'^\top}\Expect*{Z'Z'^\top}}
    &=
    \Dim\sum_{i=1}^{\Dim}\Expect*{X'X'^\top}_{ii}\Expect*{Z'Z'^\top}_{ii}\\
    &=
    \Dim\sum_{i=1}^{\Dim}\Expect*{X_i'^2}\Expect*{Z_i'^2}\\
    &\geq
    \Dim\sum_{i=1}^{\Dim}\Expect*{X_i'Z_i'}^2\\
    &=
    \left(\sum_{i=1}^{\Dim}1\right)\left(\sum_{i=1}^{\Dim}\Expect*{X_i'Z_i'}^2\right)\\
    &\geq
    \Expect*{\sum_{i=1}^{\Dim}X_i'Z_i'}^2\\
    &=
    \Expect*{X'^\top Z'}^2,
\end{align*}
where the inequalities are deduced from the Cauchy-Schwartz inequality.
\end{proof}
\begin{proof}[Proof of \cref{thm:ts-bayes-regret}]
First, observe that $(\Param,\OptimalArm_t)$ and $(\TsSample_t,\ChosenArm_t)$ are exchangeable conditional on $(\History_t,\ArmSet_t)$. Then, defining $\mu_t:=\Expect{\Param\Given\History_t,\ArmSet_t}$, we have that
\begin{align*}
    \BayesRegret(T,\PolicyTS)
    &=
    \sum_{t=1}^{T}\Expect*{\Inner{\Param,\OptimalArm_t}-\Inner{\Param,\ChosenArm_t}}\\
    &=
    \sum_{t=1}^{T}
    \Expect[\Big]{\Inner{\Param,\OptimalArm_t}}
    -
    \Expect[\Big]{\Expect*{\Inner{\Param,\ChosenArm_t}\Given\History_t}}\\
    &=
    \sum_{t=1}^{T}
    \Expect[\Big]{\Inner{\Param,\OptimalArm_t}}
    -
    \Expect[\Big]{\Inner[\big]{\Expect[\big]{\Param\Given\History_t},\Expect[\big]{\ChosenArm_t\Given\History_t}}}\\
    &=
    \sum_{t=1}^{T}
    \Expect[\Big]{\Inner{\Param,\OptimalArm_t}}
    -
    \Expect[\Big]{\Inner[\big]{\mu_t,\Expect[\big]{\OptimalArm_t\Given\History_t}}}\\
    &=
    \sum_{t=1}^{T}
    \Expect[\Big]{\Expect[\Big]
    {\Inner{\Param,\OptimalArm_t}
    -
    \Inner{\mu_t,\OptimalArm_t}\Given\History_t}}.
\end{align*}
Define $\CovMatrix_t:=\Expect[\Big]{\big(\Param-\mu_t\big)\big(\Param-\mu_t\big)^\top\Given\History_t}$. Then, it follows from \cref{lem:double-cauchy} and the independence of $\ChosenArm_t$ and $\Param$ conditional on $\HistoryPlus_t$ that
\begin{align*}
    \BayesRegret(T,\PolicyTS)
    &\leq
    \sqrt{\Dim}\sum_{t=1}^{T}\Expect[\bigg]{
    \Trace*{
    \CovMatrix_t\cdot
    \Expect[\big]{\OptimalArm_t{\OptimalArm_t}^\top\Given\History_t}
    }^{\frac12}}\\
    &=
    \sqrt{\Dim}
    \sum_{t=1}^{T}\Expect[\bigg]{
    \Trace*{
    \CovMatrix_t\cdot
    \Expect[\Big]{\ChosenArm_t{\ChosenArm_t}^\top\Given\History_t}
    }^{\frac12}}\\
    &=
    \sqrt{\Dim}\sum_{t=1}^{T}\Expect[\bigg]{
    \Trace*{
    \Expect[\Big]{
        \CovMatrix_t\cdot
        \ChosenArm_t{\ChosenArm_t}^\top
    \Given\History_t}
    }^{\frac12}}\\
    &=
    \sqrt{\Dim}\sum_{t=1}^{T}\Expect[\bigg]{
    \Expect[\Big]{
    \Trace*{
        \CovMatrix_t\cdot
        \ChosenArm_t{\ChosenArm_t}^\top
    }
    \Given\History_t}^{\frac12}
    }\\
    &=
    \sqrt{\Dim}\sum_{t=1}^{T}\Expect[\bigg]{
    \Expect[\Big]{
    {\ChosenArm_t}^\top\CovMatrix_t\ChosenArm_t
    \Given\History_t}^{\frac12}
    }\\
    &\leq
    \sqrt{\Dim}\,\sum_{t=1}^{T}\Expect[\Big]{
    {\ChosenArm_t}^\top\CovMatrix_t\ChosenArm_t
    }^{\frac12}\\
    &\leq
    \sqrt{\Dim T}\,\Expect[\Bigg]{\sum_{t=1}^{T}
    {\ChosenArm_t}^\top\CovMatrix_t\ChosenArm_t
    }^{\frac12},
\end{align*}
where the last two inequalities are obtained by applying the Cauchy-Schwartz inequality. The desired result follows from \cref{thm:stochastic-elliptical}.
\end{proof}
\end{APPENDICES}

%%%%%%%%%%%%%%%%%
\end{document}